\documentclass[letterpaper, 10 pt, conference]{ieeeconf}  

\IEEEoverridecommandlockouts                              

\title{\LARGE \bf
ZS-Puffin: Design, Modeling and Implementation of an Unmanned Aerial-Aquatic Vehicle with Amphibious Wings
}

\author{Zhenjiang Wang$^{1}$, Yunhua Jiang$^{1,2,3}$, Zikun Zhen$^{1}$, Yifan Jiang$^{1}$, Yubin Tan$^{1}$, Wubin Wang$^{1}$
\thanks{This work is supported by the National Natural Science Foundation of China under Grants No. 52371344 and No. U22A2012, Guangdong Provincial Natural Science Foundation of China under Grants No. 2024A1515012274 and 2021A1515011917, Fundamental Research Funds for the Central Universities, Sun Yat‐sen University under Grant No. 23qnpy82, and the start‐up funding to Yunhua Jiang from Sun Yat‐sen University.}
\thanks{$^{1}$School of Ocean Engineering and Technology, Sun Yat‐sen University, and Southern Marine Science and Engineering Guangdong Laboratory, Zhuhai, China.
        {\tt\small wangzhj78@mail2.sysu.edu.cn}}%
\thanks{$^{2}$Key Laboratory of Comprehensive Observation of Polar Environment (Sun Yat‐sen University), Ministry of Education, Zhuhai, China.  $^{3}$Guangdong Provincial Key Laboratory of Information Technology for Deep Water Acoustics, Zhuhai, China.
        {\tt\small Correspondence: jiangyh35@mail.sysu.edu.cn}}
}

\usepackage{romannum}
\usepackage{cuted}
\usepackage{bm}
\usepackage{cite} 
\usepackage{graphics} 
\usepackage{epsfig} 
\usepackage{mathptmx} 
\usepackage{times} 
\usepackage{amsmath} 
\usepackage{amssymb}  
\usepackage[colorlinks=true, allcolors=blue]{hyperref}
\usepackage{CJKutf8} 
\usepackage{geometry}
\begin{document}

\maketitle
\thispagestyle{empty}
\pagestyle{empty}

\begin{abstract}
Unmanned aerial-aquatic vehicles (UAAVs) can operate both in the air and underwater, giving them broad application prospects. Inspired by the dual-function wings of puffins, we propose a UAAV with amphibious wings to address the challenge posed by medium differences on the vehicle's propulsion system. The amphibious wing, redesigned based on a fixed-wing structure, features a single degree of freedom in pitch and requires no additional components. It can generate lift in the air and function as a flapping wing for propulsion underwater, reducing disturbance to marine life and making it environmentally friendly. Additionally, an artificial central pattern generator (CPG) is introduced to enhance the smoothness of the flapping motion. This paper presents the prototype, design details, and practical implementation of this concept.

\end{abstract}

\section{INTRODUCTION}
The unmanned aerial-aquatic vehicle (UAAV) is a vehicle that can fly in the air, navigate underwater, and repeatedly cross the air-water interface. It integrates the high-speed transit capabilities of unmanned aerial vehicles (UAVs) with the stealthy mobility of unmanned underwater vehicles (UUVs), thereby significantly expanding the application scenarios of single-domain vehicles. The UAAV holds broad prospects in fields such as marine resource exploration and the observation of biological behaviors. 

The UAAV needs to adapt to both air and water environments. The significant differences in the properties of these two mediums pose a considerable challenge to the propulsion system design of the UAAV. Current UAAVs can be categorized into three types: multirotor layout, fixed-wing layout, and hybrid layout. 
The multirotor layout typically possesses agile aerial maneuverability and stable water entry and exit capabilities\cite{ref:Multialzu2018Looncopter,ref:Multiravell2018modeling}. 
The fixed-wing layout is characterized by their efficient flight performance, making them suitable for long-range missions\cite{ref:fixwei2022lifting,ref:fixweisler2017testing}. 
The direct use of aerial propellers underwater results in inefficient underwater movement. To improve efficiency, some researchers have developed a rotor with dual-speed transmission or a self-folding air propeller \cite{ref:TJ-FlyingFishliu2023tj,ref:hitchhikingli2022aerial}.
The hybrid layout combines the aerial advantages of the multirotor layout and the fixed-wing layout. Additionally, these vehicles employ an additional variable buoyancy system (VBS) to achieve underwater gliding, thereby improving their submerged endurance \cite{ref:ugJFRlyu2022toward,ref:ugnezhaoelu2021design}. In prior work, we proposed an amphibious tailplane as the attitude control mechanism to reduce the structural redundancy of a hybrid UAAV\cite{ref:wang2024design}.
However, high-speed rotating propellers are not marine-life friendly, and the noise they generate can impact the marine ecosystem. Moreover, the underwater gliding mode lacks maneuverability.

Therefore, there is a need to develop a new underwater propulsion function for UAAVs. Bionic underwater robots, inspired by aquatic animals, have developed flapping-wing propulsion technology. The tuna robot, based on the body/caudal fin (BCF) propulsion mode, can achieve a maximum speed of up to four body lengths per second, making it suitable for long-duration, high-speed cruising \cite{ref:BCFzhu2019tuna,ref:BCFwang2020development}. The manta ray robot, which employs the median/paired fin (MPF) propulsion mode, exhibits better stability and maneuverability during navigation \cite{ref:MPFzhang2022novel,ref:MPFarastehfar2019relationship}. Underwater flapping-wing propulsion is safer for marine life, quieter, and offers superior stealth capabilities. The diving bird Puffin provides further inspiration, as its wings can generate lift in the air and also produce thrust by flapping underwater. Some researchers have proposed a UAAV with multi-degree-of-freedom flapping wings. These UAAVs use multiple actuators to keep the wings rigid in the air and allow them to flap underwater\cite{ref:flapgu2024bio,ref:flaphe2024novel}. Figure 1 shows the blueprint of multi-domain application scenarios for the UAAV with bionic flapping wings. In addition to basic cross-domain tasks, it is particularly adept at biological missions. Studies on flapping-wing robots, both aerial and aquatic, suggest that increasing the degrees of freedom of the wing can improve efficiency and maneuverability. However, engineering implementations must balance performance with the added complexity and cost of design. \cite{ref:freedomizraelevitz2015novel}. 
\begin{figure}[htbp]
    \centering
    \includegraphics{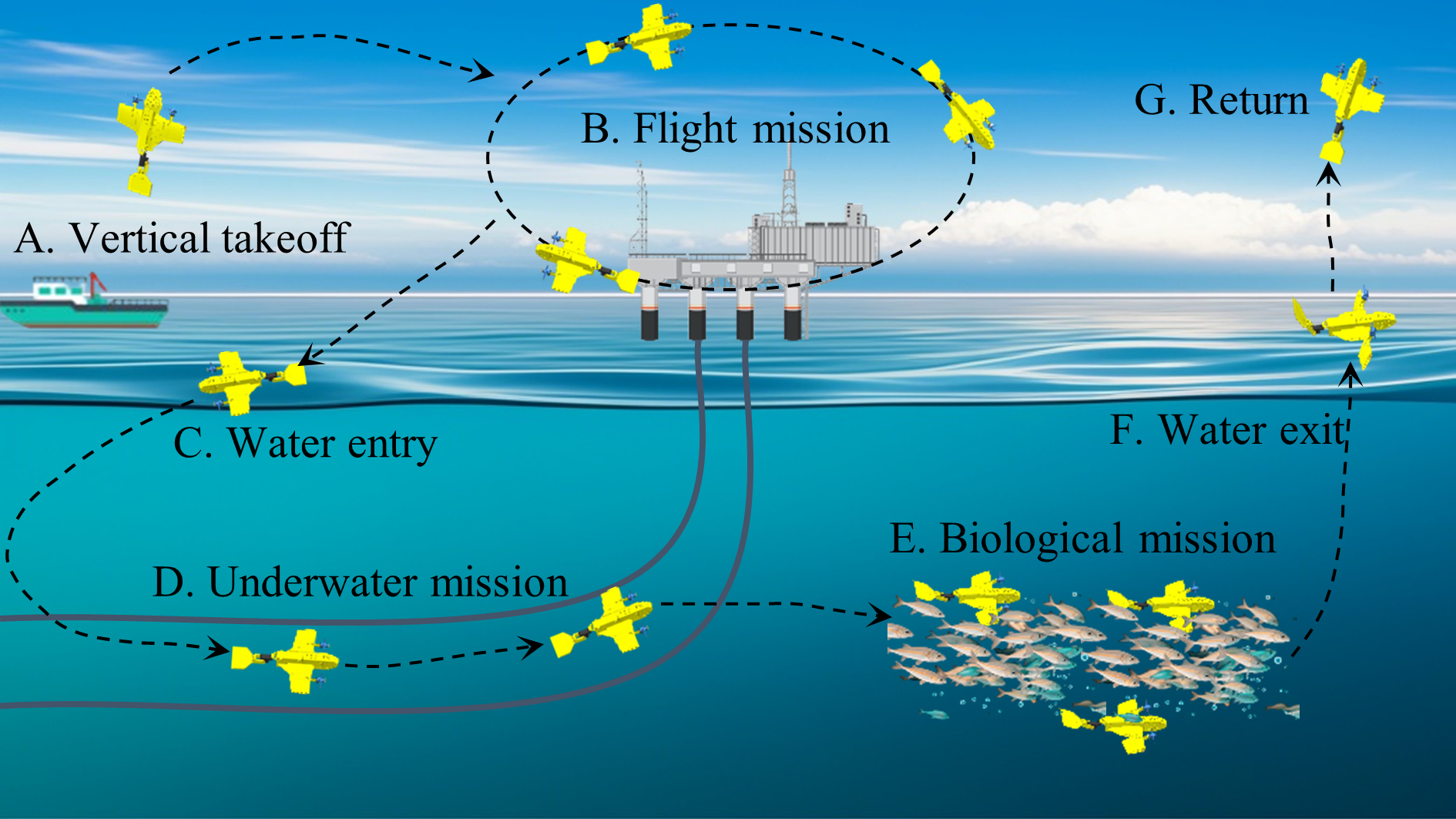}
    \caption{The blueprint of multi-domain application scenarios for the UAAV with flapping wings. (a) The vehicle takes off vertically from the platform. (b) The UAAV performs aerial missions. (c) The UAAV executes the transition from air to water. (d) The UAAV performs underwater missions. (e) The UAAV uses flapping-wing propulsion to swim alongside marine life to collect data without causing harm. (f) The UAAV executes the transition from water to air. (g) The UAAV ascends to transmit data, return to base, or proceed to the next mission site.}
    \label{fig:1}
\end{figure}

\begin{table*}[htbp]
\caption{Comparison of some HAUV prototypes}
\centering
\begin{tabular}{ccccccc}
\hline
Prototype & Type & Weight & Flight mode & Underwater thrust system & Marine biocompatibility \\
\hline
CONOPS~\cite{ref:fixweisler2017testing} & Fixed-wing & 5.6\,kg & Horizontal & Rotor & No \\
Delta-wing UAAV~\cite{ref:fixwei2022lifting} & Fixed-wing & 3\,kg & Horizontal & Rotor & No \\
Loon Copter~\cite{ref:Multialzu2018Looncopter} & Quadrotor & 2.7\,kg & Vertical & Rotor & No \\
TJ-FlyingFish~\cite{ref:TJ-FlyingFishliu2023tj} & Quadrotor & 1.63\,kg & Vertical & Tilt-rotor & No \\
Nezha \Romannum{3}~\cite{ref:ugJFRlyu2022toward} & 4+1 VTOL & 14.68\,kg & Horizontal + Vertical & VBS & No \\
Diving Hawk~\cite{ref:wang2024design} & Tilt-rotor & 8.9\,kg & Horizontal + Vertical & Tilt-rotor + VBS & No \\
ZS-Puffin & Belly-sitter & 1.61\,kg & Horizontal + Vertical & Tilt-rotor + Flipping-wing & Yes \\
\hline
\end{tabular}
\label{tab:1}
\end{table*}

A summary of the locomotion modes employed by representative UAAVs is provided in Table~\ref{tab:1}. Most of these designs exhibit limited consideration for marine biocompatibility. The primary objective of this work is to explore a novel underwater propulsion strategy for UAAVs. To achieve this, we propose the ZS-Puffin, a hybrid UAAV with amphibious wings, which combines a fixed-wing configuration with flapping-wing propulsion. The wings and tailplane on both sides have been redesigned to fully integrate with the pitch motion based flapping-wing mechanism. They are driven by servos responsible for thrust vectoring and elevator control, enabling single-degree-of-freedom periodic deflection. All wing surfaces can generate lift during horizontal flight and perform flapping propulsion underwater. The prototype uses a belly-sitter strategy for vertical takeoff, which eliminates the need for additional tail support structures. It is capable of both vertical flight and underwater vectored propulsion. Additionally, we developed a multi-mode dynamic model for the vehicle and implemented a Central Pattern Generator (CPG) control strategy to achieve smoother underwater flapping wing propulsion.

\section{PROTOTYPE VEHICLE DESIGN}

\subsection{Configuration Overview}
\begin{figure*}[htbp]
    \centering
    \includegraphics{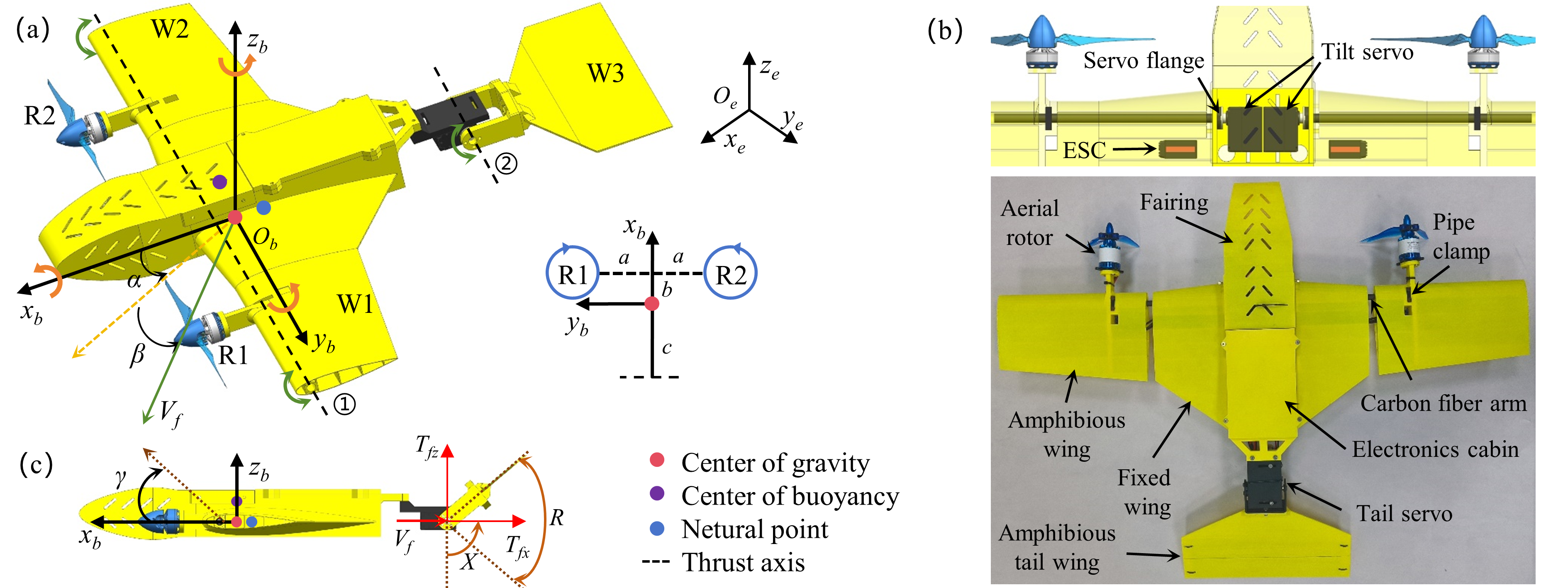}
    \caption{(a) The geometric model of the ZS-Puffin with coordinate system and relevant parameter markings. (b) The prototype of ZS-Puffin and the labels of its main components. (c) The side view of ZS-Puffin with parameter markings.}
    \label{fig:2}
\end{figure*}

ZS-Puffin adopts a thrust-vectored belly-sitter framework. As shown in Fig. \ref{fig:2}, it has a wingspan of 690 mm, a length of 580 mm, and a height of 50 mm. The prototype weighs 1.61 kg. The front part of the fuselage is a fairing designed to optimize fluid dynamic performance, while the rear part is a watertight electronics cabin that houses all onboard electronic components. On the outer side of the fuselage are two amphibious wings, which lack separate ailerons. These wings, driven by tilting servos, can rotate around an internal carbon fiber arm and integrate the functions of fixed wings, ailerons, and flapping wings. The air rotors are mounted on the amphibious wings and tilt synchronously with them. To increase the wing area, two fixed wings are attached to the inner side of the fuselage. The rear of the prototype features an amphibious tail wing, which combines the functions of the tailplane, elevator, and flapping wing. It is driven by a tail servo.
Through weight distribution, the vehicle's center of gravity ($C_G$) is positioned between thrust axis 1 (carbon fiber arm) and the neutral point. As shown in Fig. \ref{fig:2}(a), it is aligned parallel to R1 and R2. This arrangement ensures weight balance for vertical flight and longitudinal static stability during horizontal flight. The vehicle has slight positive buoyancy, with its center of buoyancy ($C_B$) vertically positioned above the $C_G$ to allow stable floating on the water surface.

\begin{figure}[htbp]
    \centering
    \includegraphics{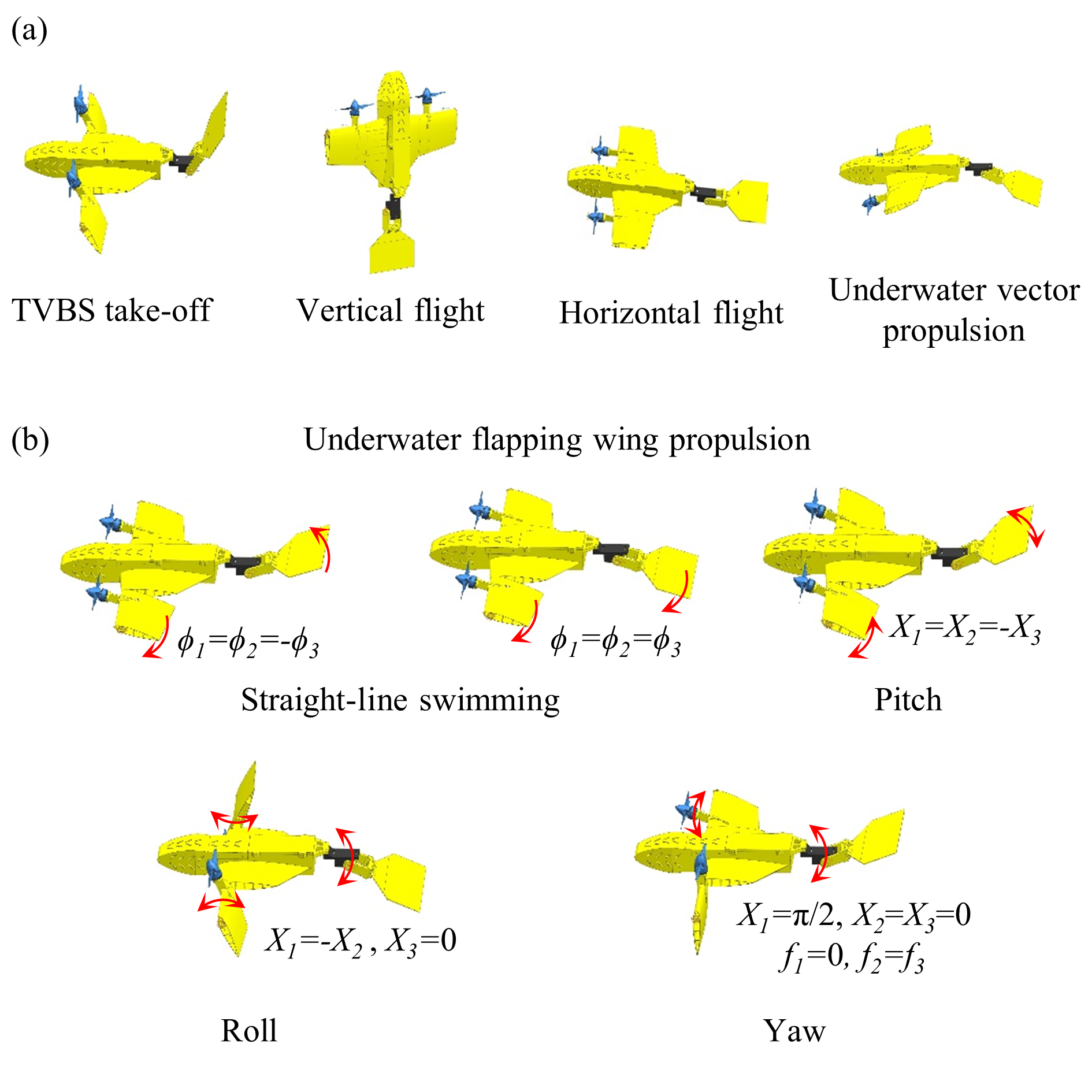}
    \caption{Schematic diagram of the operational modes of the ZS-Puffin. (a) The flight mode and underwater vector propulsion mode. (b) The underwater flapping-wing propulsion mode and main control parameters.}
    \label{fig:3}
\end{figure}

Fig. \ref{fig:3} illustrates the various operational modes of ZS-Puffin. The thrust-vectored belly-sitter strategy allows the vehicle to land on its belly with two rotors facing upward, unlike tail-sitter configurations. This design enables vertical takeoff without the need for additional tail supports. Vertical and horizontal flight modes are similar to those of conventional tail-sitter drones \cite{ref:lovell2023attitude}. Computational fluid dynamics results indicate that the vehicle can generate sufficient lift to balance its weight during horizontal flight. This performance is achieved using a NACA 0015 airfoil with an area of 0.076 m², at an angle of attack of 10 degrees and a speed of 18.6 m/s. The underwater vectored propulsion mode uses the air rotors to provide thrust underwater. The control moments are generated by directly altering the thrust and wing direction via servos. Considering the inefficiency of air rotors in water, this mode is suitable for agile maneuvers over short underwater distances. The underwater flapping-wing propulsion mode is shown in Fig. \ref{fig:2}(b). By adjusting the control parameters of the CPG strategy, the vehicle's flapping-wing motion can be smoothly transitioned. Further details will be provided in Section \Romannum{3} B.

\subsection{Avionics}
\begin{figure}[htbp]
    \centering
    \includegraphics{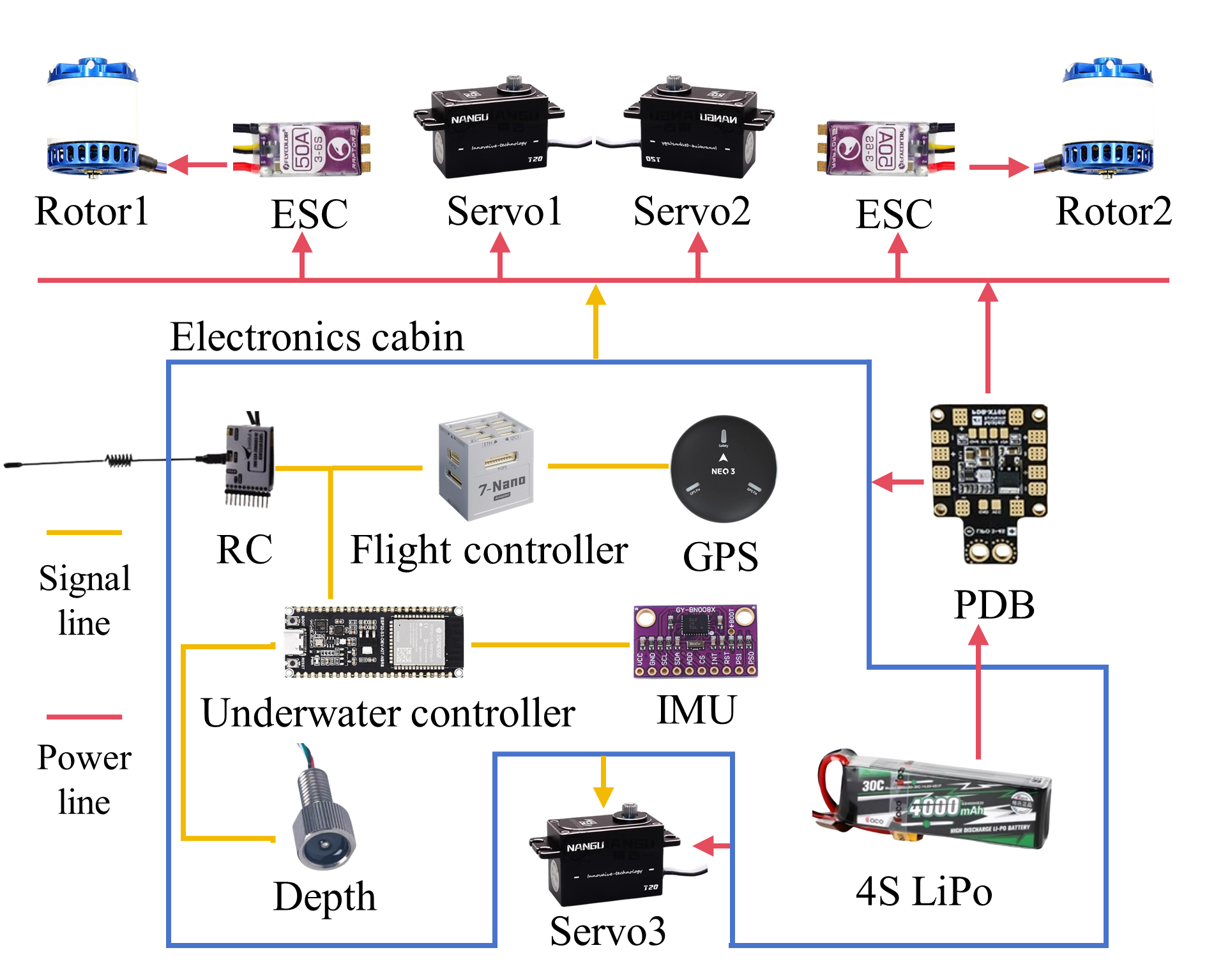}
    \caption{Electronic and hardware framework of ZS-Puffin.}
    \label{fig:4}
\end{figure}

The avionics system of the prototype is shown in Fig. \ref{fig:4}. A CUAV 7 nano flight controller with a GPS module is used to control and switch flight modes, as well as to record the vehicle's position in the air. An ESP 32 underwater controller is connected to an inertial measurement unit (IMU) and a depth sensor. It executes preprogrammed control algorithms underwater and senses the vehicle's attitude and depth. The controllers are also connected to a 433 MHz receiver, enabling reception of ground commands in shallow water. All water-sensitive components are housed in a watertight electronics cabin to prevent water ingress. Two 2216 2600 KV air rotors are driven by two 50 A electronic speed controllers (ESC). These rotors can generate a maximum thrust of 31.6 N in air. The ESCs are coated with silicone sealant for waterproofing and are directly immersed in water to enhance heat dissipation. Three waterproof servos are used to enable direct underwater operation. A 4 S Li‐Po battery powers the electronics compartment and servos at 5 V through a power distribution board (PDB). Additionally, it provides full battery voltage to the two ESCs.

\section{DYNAMICS AND CONTROL}

\subsection{Multi-Modal Dynamics}
The simplified geometric model of ZS-Puffin with coordinate system annotations and the positive direction is shown in Fig. \ref{fig:2}(a). Two frames are defined: a right-hand inertial frame $\{\bm{E}\}(O_e, x_e, y_e, z_e)$. Here, $O_e$ is the origin fixed to the water surface. The positive direction of the x-axis points towards the vehicle's nose, and the z-axis is vertically oriented upwards relative to the Earth. $\{\bm{B}\}(O_b,x_b,y_b,z_b)$ stands for the body fixed frame of the vehicle, where $O_b$ is fixed on the center of mass ($C_G$) of the ZS-Puffin. The position vector in $\{\bm{E}\}$ is defined as $\bm{P} = [x\ y\ z]^T$, and the attitude vector is $\boldsymbol{\Theta} = [\phi\ \theta\ \psi]^T$. The velocity vector and angular rate vector in $\{\bm{B}\}$ are $\boldsymbol{V} = [u\ v\ w]^T$ and $\boldsymbol{\Omega} = [p\ q\ r]^T$ respectively. 

In traditional aircraft modeling, the attitude of an airplane is typically described using ZYX Euler angles \cite{ref:nelson1998flight} . Although ZYX Euler angles encounter a singularity at a pitch angle of 90 degrees, this issue is generally negligible for conventional aircraft, as they seldom operate at such extreme pitch angles. However, during mode transition, tail-sitter and belly-sitter UAVs often reach pitch angles close to 90 degrees. At this point, the ZYX Euler angles are no longer suitable. Using quaternions to describe the vehicle's attitude can resolve the singularity issue, but this method lacks the intuitiveness.

In this paper, we employ the ZXY Euler angles to represent the vehicle's attitude. ZXY Euler angles begin by aligning the body frame with the inertial frame.  The body-fixed frame then rotates around $z_ b$ by angle $\psi$,  followed by a rotation around the new $x_b$  by angle $\phi$, and finally around the new $y_b$  by angle $\theta$  \cite{ref:2017Modeling,ref:ugnezhaoelu2021design}. The ZXY Euler angles switch the order of the last two rotations. This modification does not affect intuitiveness but shifts the singularity from $\theta = \pm \frac{\pi}{2}$ to $\phi = \pm \frac{\pi}{2}$, a point that the ZS-Puffin rarely reaches. The resulting rotation matrix is $\boldsymbol{R_B^E}$, represents the transformation from $\{\bm{B}\}$ to $\{\bm{E}\}$ and is given by: 
\begin{align}
    \boldsymbol{R_B^E} =
\begin{bmatrix}
c\theta c\psi - s\phi s\theta s\psi & -c\phi s\psi & s\theta c\psi + s\phi c\theta s\psi \\
c\theta s\psi + s\phi s\theta c\psi & c\phi c\psi & s\theta s\psi - s\phi c\theta c\psi \\
-c\phi s\theta & s\phi & c\phi c\theta
\end{bmatrix} \label{Equ:REB}
\end{align}
Notice that the $\bm{R}_B^E$ is an orthogonal matrix, which means that $\bm{R}_B^{E^{-1}} = (\bm{R}_B^E)^T = \bm{R}_E^B$ . The transformation matrix of angular rate $\bm{W}$ is given by:
\begin{align}
    \bm{W} = 
\begin{bmatrix}
c\theta & 0 & s\theta \\
s\theta \ t\phi & 1 & -c\theta \ t\phi \\
-s\theta/c\phi & 0 & c\theta/c\phi
\end{bmatrix}
\end{align}
where $c(\cdot)$, $s(\cdot)$ and $t(\cdot)$ represent the abbreviation of $\cos(\cdot)$, $\sin(\cdot)$ and $\tan(\cdot)$, respectively. Then the kinematic equation of the vehicle can be expressed as:
\begin{align}
\dot{\bm{P}} = \bm{R}_B^E \bm{V},\quad \dot{\bm{\Theta}} = \bm{W} \bm{\Omega} \label{eq:kinematic}.
\end{align}
The multi-modal dynamics for the vehicle under $\bm{B}$ are given based on the Newton-Euler equation, as follows:
\begin{equation}
\begin{split}
    (\bm{M }_0- k\bm{M}_a)\bm{\dot{v}} + [(\bm{M }_0- k\bm{M}_a)\bm{\Omega}] \bm{\times v} 
    +  \bm{F} _f+   \bm{F}_r = \bm{F}_j \\
\end{split}
\end{equation}
\begin{equation}
   \begin{aligned}
    (\bm{J} _0- k\bm{J}_a)\bm{\dot{\Omega}} + (k\bm{M}_a\bm{v}) \times \bm{v} - [(\bm{J}_0 - k\bm{J}_a)\bm{\Omega}] \times \bm{\Omega} \\ +  \bm{M}_f + k\bm{M}_r = \bm{M}_j, \ j = a(air), w(water)
\end{aligned} 
\end{equation}
where $\bm{F}_j$ and $\bm{M}_j$ represent the control force and control torque in water or air, respectively. These are generated by two rotors (R1 and R2) and three amphibious wings in flapping-wing mode (W1, W2, and W3). The $k$ denotes the state flag, where $k = 0$ and $1$ represent aerial and aquatic states respectively.

\textbf{1) Added Mass and Inertia:} $m$ is the mass of vehicle and $\bm{M}_0 = diag[m\ m\ m]$. $\bm{J}_0 = diag[I_{xx}\ I_{yy}\ I_{zz}]$ is the moment of inertia around the $x_b$, $y_b$, and $z_b$. $\bm{M}_a = diag[X_{\dot{u}},Y_{\dot{v}},Z_{\dot{w}}]$, $\bm{J}_a = {diag}[K_{\dot{p}},M_{\dot{q}},N_{\dot{r}}]$ represent the added mass and inertia matrices, respectively, for the 6 degrees of freedom (6DOF) in hydrodynamics \cite{ref:fossen2011handbook}.

\textbf{2) Fluid Force and Moment:} $\bm{F} _f \text{ and } \bm{M}_f$ are the fluid force and moment of the vehicle in air or water, which can be expressed as \cite{ref:etkin1995dynamics,ref:2017Modeling}: 
\begin{equation}
\begin{split}
    \bm{F} _f= 
\begin{bmatrix}
c\alpha c\beta & -c\alpha s\beta & -s\alpha \\
s\beta  & c\beta & 0 \\
s\alpha c\beta & -s\alpha s\beta & c\alpha
\end{bmatrix}
\begin{bmatrix}
0.5\rho_jV_f^2SC_D \\
0.5\rho_jV_f^2SC_Y \\
0.5\rho_jV_f^2SC_L
\end{bmatrix}, \\
 \bm{M} _f= 
\begin{bmatrix}
0.5\rho_jV_f^2S\overline{c}C_l \\
0.5\rho_jV_f^2S\overline{c}C_m \\
0.5\rho_jV_f^2S\overline{c}C_n
\end{bmatrix}, j = a(air), w(water)  
\end{split}
\end{equation}
where $\rho_j$ is the density of air or water, $S$ is the wing area, $\overline{c}$ is the characteristic length, and $V_f$ is fluid speed. $\alpha \text{ and } \beta$ are the angle of attack and sideslip angle, respectively, as shown in Fig. \ref{fig:2}(a). $C_L$, $C_D$ and $C_Y$ are respectively the lift, drag and side force coefficients, and $C_l$, $C_m$ and $C_n$ are the rolling, pitching and yawing coefficients, respectively. These coefficients are in general functions of fluid speed $V_f$ , angle of attack $\alpha$, and sideslip angle $\beta$. 

\textbf{3) Restoring Force and Moment:} The gravity vector of vehicle can be expressed as $\bm{f_g} = [0\ 0\ -mg]^T$, and the buoyancy vector is $\bm{f_b} = [0\ 0\ \rho_wgV_{vol}]^T$. Here, $g$ is the acceleration of the gravity, and $V_{vol}$ is the volume of prototype. Additionally, when the vehicle is fully submerged, the coordinates of the center of buoyancy ($C_B$) in the body frame $\{\bm{B}\}$ are denoted as $\bm{r}_B = [x_v\ y_v\ z_v]^T$. The restoring force $\bm{F}_r$ and moment $\bm{M}_r$ in the body frame are given by:
\begin{equation}
    \bm{F}_r = \bm{R}_E^B(\bm{f} _b+ \bm{f}_g),\ \bm{M}_r = \bm{r}_b \times \bm{R}_E^B \bm{f}_b
\end{equation}
\textbf{4) Control Force and Moment:} R1 and R2 can generate thrust in both air and water. The thrust model of the rotor is described as follows:
\begin{equation}
    \begin{split}
        T_{rji} = C_{Tj} \omega_{ji}^2,\ M_{rji} = C_{Qj} T_{rji}, \ i = 1, 2 
    \end{split}
\end{equation}
where, $T_{rji}$ and $M_{rji}$ are the thrust and reaction torque generated by two rotors in air or water, respectively. $C_{Tj}$ is the thrust coefficient related to the rotor parameters, $C_{Qj}$ is the reaction torque coefficients, and $\omega_{ji}$ is the rotor speed. The position of R1 and R2 in the coordinate system are marked in Fig. \ref{fig:2}(a). $b$ and $a$ are respectively the vertical and horizontal distances from the $C_G$, while $\gamma_1$ and $\gamma_2$ are the tilt angles (in Fig. \ref{fig:2}c) of R1 and R2. The two tilt-rotors operate in three modes: vertical flight, horizontal flight, and underwater vector propulsion. They generate pitch and roll moments through tilting, while the yaw moment is produced by the speed differential between the two rotors. The models are as follows:
\begin{align}
\bm{F}_j = \begin{bmatrix}
T_1c\gamma_1 + T_2c\gamma_2 \\
0 \\
T_1s\gamma_1 + T_2s\gamma_2
\end{bmatrix},\
\bm{M}_j = 
\begin{bmatrix}
    T_1as\gamma_1 - T_2as\gamma_2\\
    -T_1bs\gamma_1 - T_2bs\gamma_2\\
    T_2ac\gamma_2 - T_1c\gamma_1
\end{bmatrix}
\end{align}
W1 to W3 generate thrust in water through periodic flapping motions. The dynamic models of the flapping wing performing pitching motion underwater are described as follows \cite{ref:liu2020effects,ref:alam2020dynamics}:
\begin{equation}
\begin{split}
   T_{fxi} &= 0.5 \rho_w (s(X_i-\alpha)V_f)^2 S\overline{C} _{fx},\\  T_{fzi} &= 0.5 \rho_w (s(X_i-\alpha)V_f)^2 S\overline{C} _{fz},\ i = 1,\ 2, \ 3
\end{split}
\end{equation}
where, $ T_{fxi}$ is the trust, $T_{fyi}$ is the vertical force, and $X_i$ is the offset of wing $i$, as shown in Fig. \ref{fig:2}(c). The coefficients $\overline{C} _{fx}$ and $\overline{C} _{fy}$ are the time-averaged coefficients of the corresponding forces over one period $T$, respectively. These coefficients are defined as follows:
\begin{equation}
\begin{split}
\overline{C}_{fx} &= -\frac{1}{nT} \int_{t_0}^{t_0+nT} \frac{T_{fx}(t)}{0.5 \rho_w V_f^2 S} dt,
\\ \overline{C}_{fz} &= -\frac{1}{nT} \int_{t_0}^{t_0+nT} \frac{T_{fz}(t)}{0.5 \rho_w V_f^2 S}dt
\end{split}
\end{equation}
In this study, the forces generated by the flapping wing result from an oscillatory pitching motion. As a result, the resultant force of $T_{fz}$ over one period is zero, leading to $\overline{C}_{fz}=0$. $c$ represents the vertical distance from the $C_G$ to the thrust axis 2 (the axis of action of W3).  The models of underwater flapping-wing are as follows: 
\begin{align}
    \bm{F}_w &= 
\begin{bmatrix}
T_{fx1}sX_1 + T_{fx2}sX_2+T_{fx3}sX_3 \\
0 \\
T_{fx1}cX_1+T_{fx2}cX_2+T_{fx3}cX_3
\end{bmatrix}
\end{align}
\begin{align}
     \bm{M}_w &= 
\begin{bmatrix}
    T_{fx1}acX_1-T_{fx2}acX_2\\
    T_{fx3}ccX_3-T_{fx1}bcX_1-T_{fx2}bcX_2\\
    T_{fx2}sX_2-T_{fx1}sX_1
\end{bmatrix}
\end{align}
\subsection{Control Strategies}
\begin{figure} [htbp]
    \centering
    \includegraphics{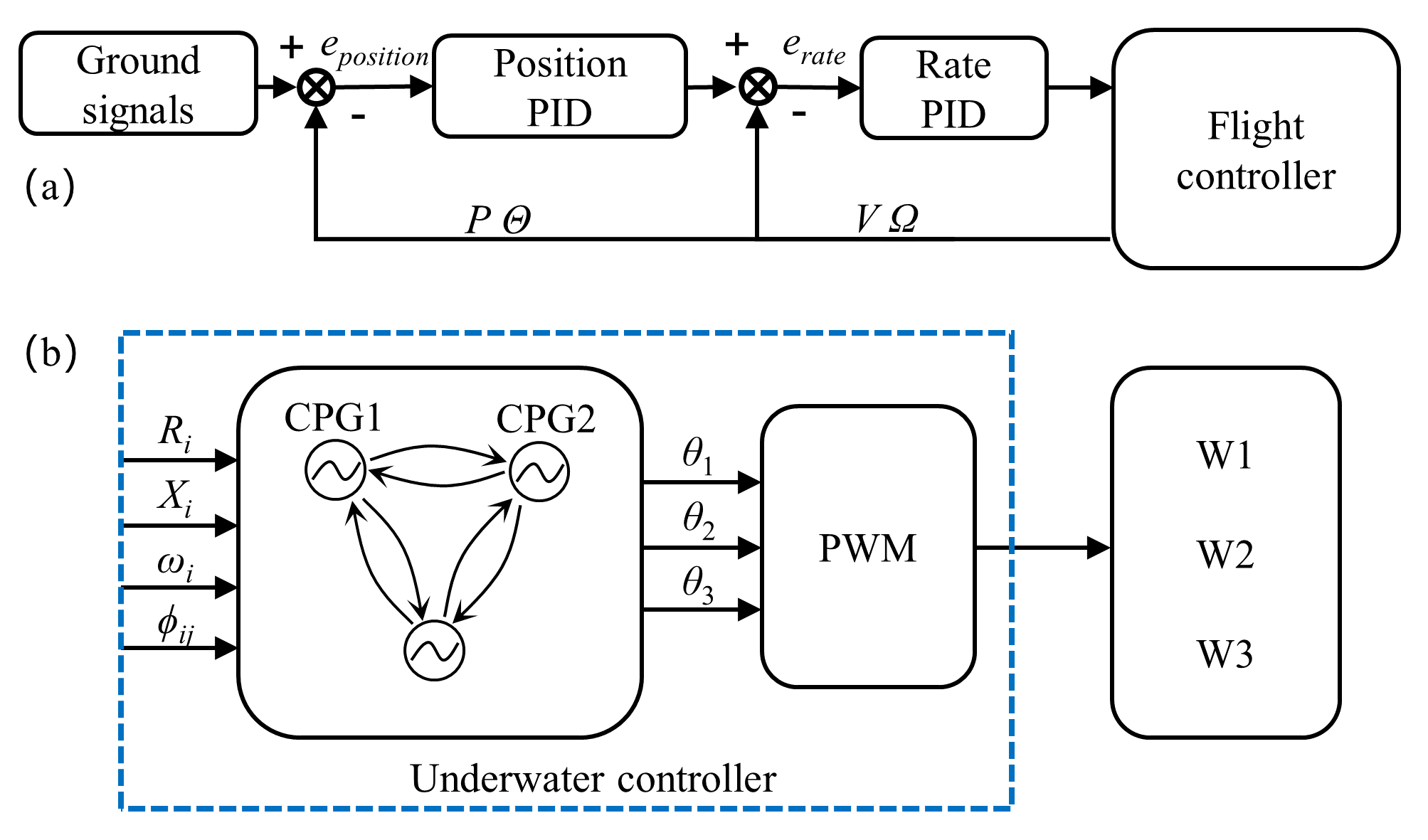}
    \caption{The control strategy block diagram of ZS-Puffin. (a) Flight mode. (b) Underwater flapping-wing mode.}
    \label{fig:5}
\end{figure}

A cascaded closed-loop PID control strategy is employed for the operation of ZS-Puffin in both vertical and horizontal flight modes, as shown in Fig. \ref{fig:5}(a). The first stage tracks the vehicle's position, while the second stage tracks its velocity. The output of the first stage serves as the input to the second stage. The vehicle’s stability in water is enhanced by moving the battery to lower the $C_G$ beneath the $C_B$. As a result, an open-loop control strategy is employed for underwater vectored propulsion. The pilot directly controls the rotors and servos via a mixing control module.

As for the underwater flapping-wing mode, our locomotion controller is based on an artificial central pattern generators (CPG) model. CPGs offer an effective approach for controlling robotic fish equipped with multiple actuators to generate periodic motions. CPG can produce coordinated, high-dimensional rhythmic output signals using only simple, low-dimensional input signals. Moreover, the limit cycle inherent in CPGs ensures that the control system can smoothly return to stable states after external disturbances. While the traditional sine-based method can represent control signals for periodic flapping motions, it fails to eliminate discontinuities when the flapping mode changes \cite{ref:crespi2008controlling,ref:cao2015applying}. Our CPG controller is implemented as a system of three coupled, amplitude-controlled phase oscillators. Each wing is assigned to one oscillator, as shown in Fig. \ref{fig:5}(b). The implementation of oscillator i is described as follows:
\begin{align}
\dot{\phi}_i &= 2\pi f_i + \sum_{j} (w_{ij} \sin(\phi_j - \phi_i - \phi_{ij}))\label{equ:phi}, \\
\ddot{r}_i &= a_r \left( \frac{a_r}{4}(R_i - r_i) - \dot{r}_i \right)\label{equ:r}, \\
\ddot{x}_i &= a_x \left( \frac{a_x}{4}(X_i - x_i) - \dot{x}_i \right)\label{equ:x}, \\
\theta_i &= x_i + r_i \cos(\phi_i)\label{equ:theta}
\end{align}
where $\theta_i$ is output of  the oscillator $i$ (in radians). The state variables $\phi_i$, $r_i$, and $x_i$ represent the phase, amplitude, and offset of the oscillations, respectively (in radians). The desired frequency $f_i$ and desired amplitude $R_i$ are power control parameters, while the desired offset $X_i$ is direction control parameter. The thrust magnitude and direction of W1 to W3 can be flexibly adjusted by modulating these control parameters, enabling precise control of the vehicle's motion. Equ.\ref{equ:r} and \ref{equ:x} are second-order linear differential equations. It is easy to demonstrate that, starting from arbitrary initial conditions, the state variables $r_i$ and $x_i$ asymptotically converge to $R_1$ and $X_i$, respectively. The parameters $a_r$ and $a_x$ are constant positive gains that influence the convergence rate and smoothness of the system. The coupling weights $w_{ij}$ determine the rate of phase evolution. $\phi_{ij}$ are phase biases which determine how oscillator $j$ influences oscillator $i$. 

\begin{figure}[htbp]
    \centering
    \includegraphics{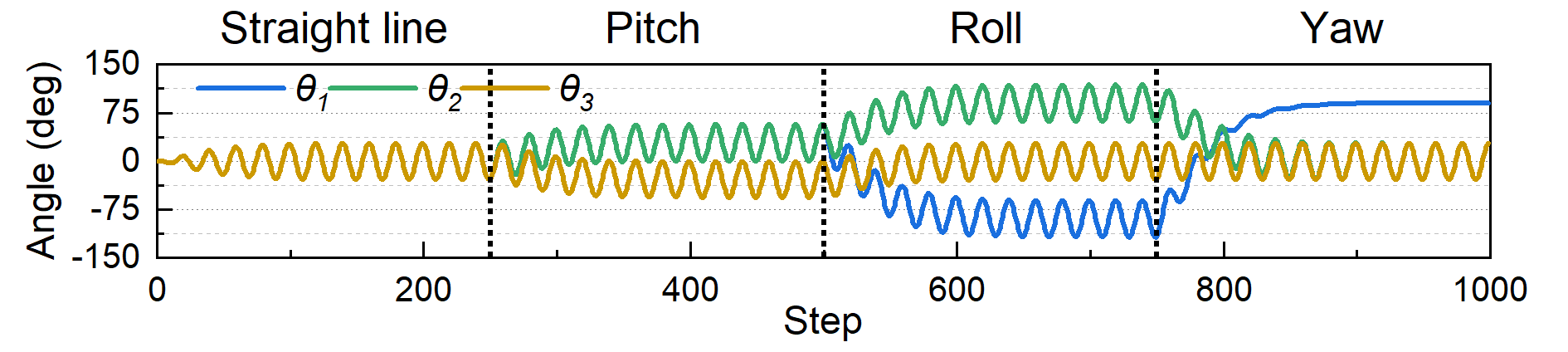}
    \caption{The output of oscillators.}
    \label{fig:6}
\end{figure}

CPG1 to CPG3 correspond to the W1 - W3, respectively. Different flapping behaviors can be achieved by modulating the CPG control parameters. Examples of flapping-wing propulsion mode are illustrated in Fig. \ref{fig:3}(b). The typical control parameters for flapping-wing behavior are listed in Table \ref{tab:2}. The output of oscillator $\theta_i$ is shown in Fig. \ref{fig:6}. As can be seen, when the flapping mode is switched, the output remains smooth. For all these behaviors, the speed of locomotion can be adjusted by varying the frequency $f_i$ and the amplitudes $R_i$ of oscillator $i$. Typically, the speed of locomotion increases with these parameters until the torque limits of the motors are reached.

\begin{table}[htbp]
\centering
\caption{The control parameters of flapping behaviors}
\begin{tabular}{|c|c|}
\hline
Forward straight-line swimming& $X_i=0, \phi_1=\phi_2=\pm\phi_3$ \\ \hline
Roll& $X_1=-X_2, X_3=0$\\ \hline
Pitch& $X_1=X_2=-X_3$\\ \hline
Yaw & $X_{1 \ or \ 2}=\pi/2, X_{2\ or\ 1}=X_3=0$\\ \hline
\end{tabular}
\label{tab:2}
\end{table}

\section{EXPERIMENTAL RESULT}
\begin{figure}[htbp]
    \centering
    \includegraphics{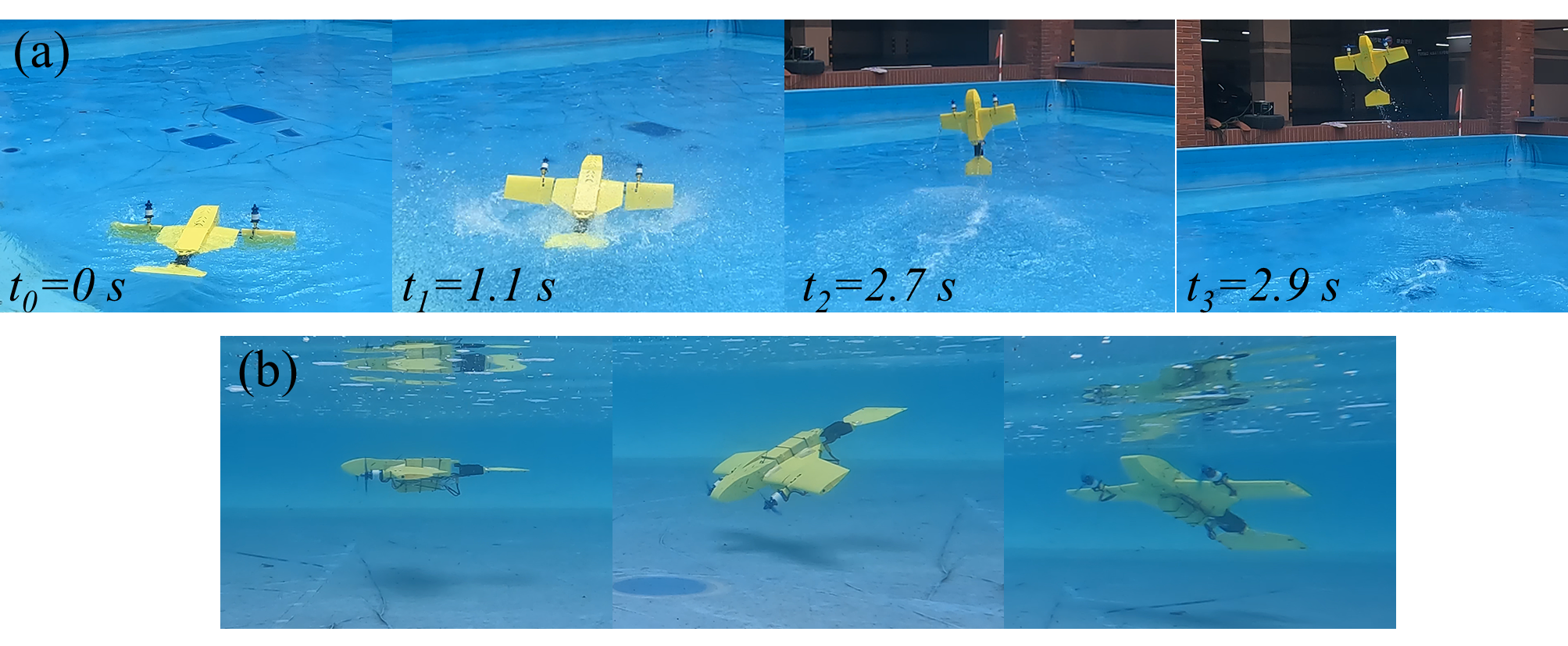}
    \caption{(a) The water exit test under the belly-sitter configuration. (b) The test of underwater vectored propulsion. }
    \label{fig:7}
\end{figure}

The water exit performance of the prototype was tested, as shown in Fig. \ref{fig:7}(a).  The vehicle floats stably on the water surface in a belly-sitter configuration. At $t_0=0$ s, the tilting angles of W1 and W2 are maintained at the extreme positions. This ensure that R1 and R2 are as far away from the water surface as possible. At  $t_1=1.1$ s, as the thrust increases, the vehicle's fuselage begins to exit the water. At  $t_2=2.7$ s, the vehicle has fully exited the water. Simultaneously, the water trapped in the structural cavities of the prototype gradually drains out. At  $t_3=2.9$ s, the prototype completes its water exit operation and resumes vertical flight. The entire water-exit process takes approximately 3 s, achieved without the need for any additional tail support mechanisms. 

The underwater vectored propulsion operation of the vehicle is shown in Fig. \ref{fig:7}(b). Although using air propellers directly underwater is inefficient, this mode is acceptable as a trade-off. This is due to the reduction in payload and the increase in structural complexity that would result from adding extra underwater actuators. The results demonstrate that the vectored thrust generated by R1 and R2 under servo operation, combined with the control moments produced by the deflection of W1 to W3, effectively maneuver the vehicle underwater. By timing five fixed-distance movements, the average speed of the vehicle in the underwater vectored propulsion mode is 0.63 m/s.

\begin{figure}[htbp]
    \centering
    \includegraphics{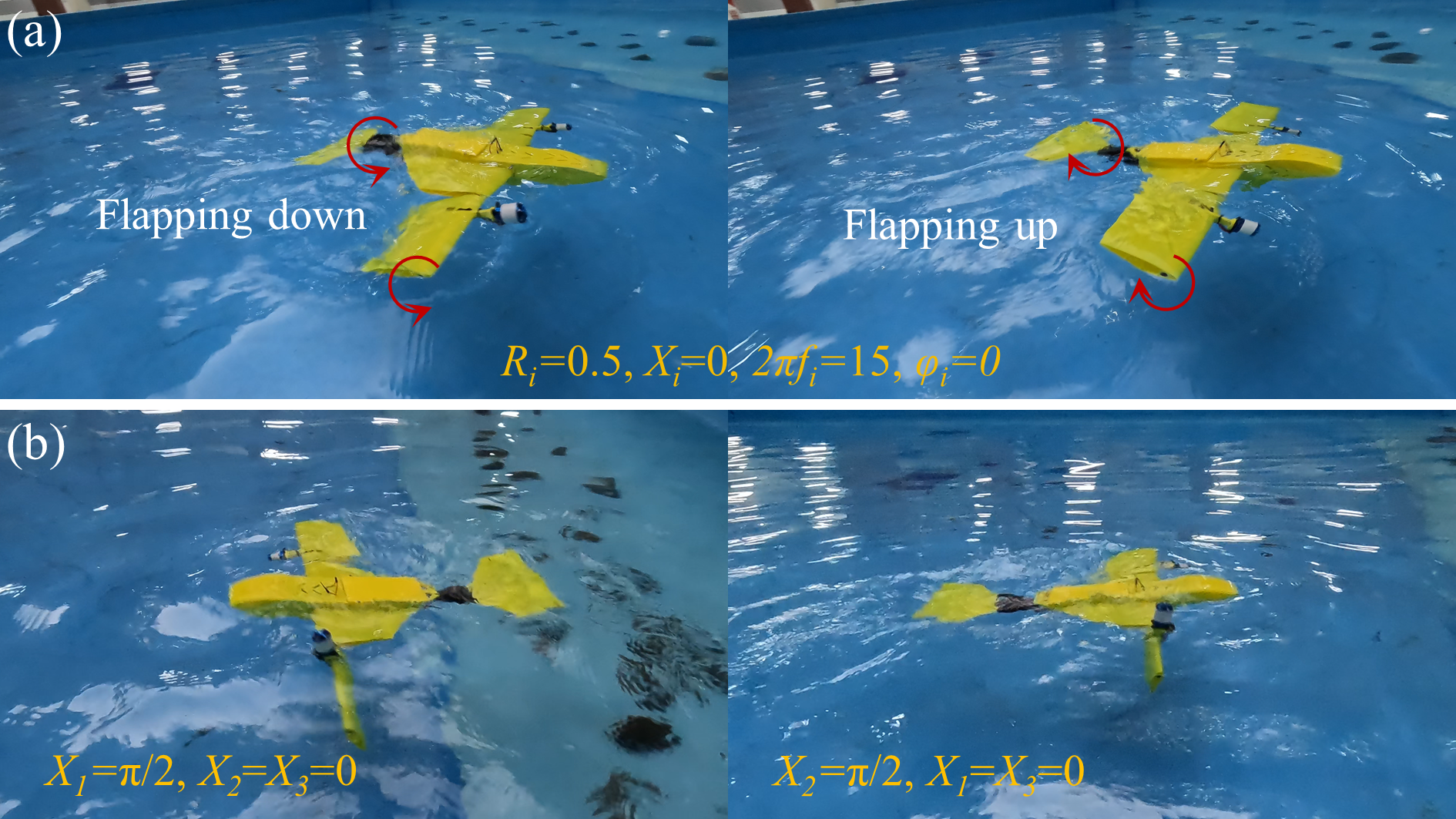}
    \caption{The test of underwater flapping-wing propulsion mode. (a) Forward straight-line swimming test. (b) Positive yaw and negative yaw tests. }
    \label{fig:8}
\end{figure}

In the tests of the underwater flapping-wing propulsion mode, the maximum torque of the servos was utilized. The control parameters were set as \( 2\pi f_i = 15 \), \( R_i = 0.5 \), \( X_i =  0 \), and \( \phi_1 = \phi_2 = \phi_3  \) (indicating no phase difference between W1 to W3). The straight-line navigation and yaw operations are shown in Fig. \ref{fig:8}. The thrust generated by the periodic flapping of W1 to W3 allowed the vehicle to move stably through the water. Moreover, by adjusting the control parameters, the vehicle is capable of performing turning maneuvers. The onboard storage recorded the vehicle's attitude data during the tests, as shown in Fig. \ref{fig:7}. During all the flapping-wing propulsion tests, the vehicle's attitude experienced regular fluctuations due to the transient vertical forces generated by the flapping wings. Although the resultant vertical force over one period is zero, the periodic vertical forces continuously disturb the vehicle's attitude during flapping-wing propulsion. In the straight-line navigation test, two different phase differences were employed. In Test 1, \( \phi_1 = \phi_2 = \phi_3 \), indicating no phase difference between W1, W2, and W3. In Test 2, \( \phi_1 = \phi_2 = -\phi_3 \), indicating a 180 degree phase difference between W1 and W2 relative to W3. 

\begin{figure}[htpb]
    \centering
    \includegraphics{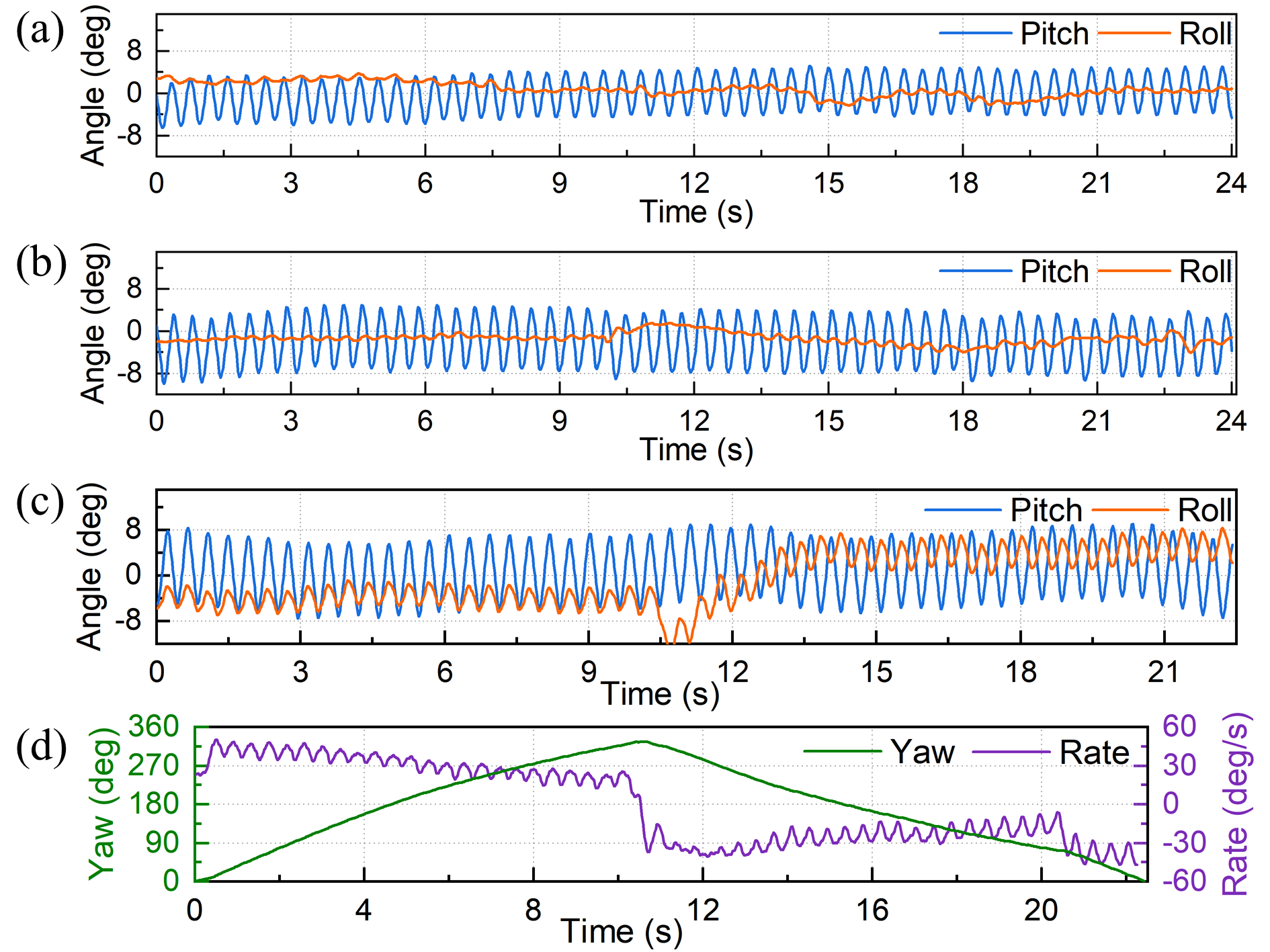}
    \caption{The attitude data of underwater flapping-wing propulsion mode tests. (a) Test 1 and (b) test 2 are straight-line navigation results. (c) and (d) are yaw results of test 3.}
    \label{fig:9}
\end{figure}

The pitch and roll data are shown in Fig. \ref{fig:9}(a) and (b). The pitch angle fluctuation range in Test 1 is smaller than that in Test 2. This occurs because the vehicle's $C_G$ is located between thrust axis 1 and thrust axis 2. When \( \phi_1 = \phi_2 = \phi_3 \), the vertical forces on the thrust axis 1 and thrust axis 2 act in the same direction. However, their magnitudes differ due to the varying wing areas, generating a periodic pitching moment around the $y_b$. This moment causes the pitch angle fluctuations observed in Fig. \ref{fig:9}(a). In contrast, when \( \phi_1 = \phi_2 = -\phi_3 \), the vertical forces act in opposite directions, resulting in a larger pitching moment. This larger moment leads to a greater pitch angle fluctuation range, as shown in Fig. \ref{fig:9}(b).  These pitch fluctuations can negatively impact the vehicle's flapping-wing propulsion performance. By increasing the vehicle's projected area in the $x_by_b$-plane, these fluctuations augment the drag in the $x_b$ direction. The average speed in Test 1 is 0.29 m/s, while in Test 2 it is 0.21 m/s. The vehicle achieves higher speeds when pitch fluctuations are minimized. The roll angle fluctuations around 0 degrees are primarily caused by water surface waves. As a result, the fluctuation ranges are similar for both Test 1 and Test 2.

Test 3 is a yaw test utilizing flapping-wing propulsion. Positive yaw is achieved when $X_1 =\pi /2$, and negative yaw is achieved when $X_2 =\pi /2$. When $X_i =\pi /2$, W$i$ is perpendicular to the $x_b$-$y_b$ plane and fully submerged in water. As shown in Fig. \ref{fig:9}(c), the vehicle's pitch angle fluctuates during yaw operations due to the influence of vertical forces. During positive yaw, the roll angle fluctuates below 0 degrees, while during negative yaw, it fluctuates above 0 degrees. This is because the offset of a single wing changes the vehicle's buoyancy distribution, shifting the $C_B$ to the opposite side. Fig. \ref{fig:9}(d) shows that the vehicle's yaw rate also exhibits oscillatory behavior. The peak yaw rate occurs at a frequency of approximately 2.4 Hz, which matches the flapping frequency. Moreover, similar oscillatory phenomena are observed in both the pitch and roll angles.

\section{CONCLUSIONS}
This paper presents an unmanned aerial-aquatic vehicle named ZS-Puffin, which is equipped with amphibious wings. The prototype features multiple operational modes, and a multi-mode dynamic model of the vehicle has been developed. The feasibility, functionality, and maneuverability of the prototype have been experimentally validated. The amphibious wing, designed with a single degree of freedom in pitch, features a simple structure. It can provide lift in the air and generate thrust underwater through flapping motions. The enhance the vehicle's adaptability for marine biological tasks. The implementation of an artificial central pattern generator (CPG) mode significantly improves the flexibility and smoothness of the underwater flapping propulsion. Future research will focus on developing advanced control algorithms to enhance the efficiency of the flapping mode and further reduce pitch oscillations.
\bibliographystyle{IEEEtran}
\bibliography{references}

\begin{thebibliography}{10}
\providecommand{\url}[1]{#1}
\csname url@rmstyle\endcsname
\providecommand{\newblock}{\relax}
\providecommand{\bibinfo}[2]{#2}
\providecommand\BIBentrySTDinterwordspacing{\spaceskip=0pt\relax}
\providecommand\BIBentryALTinterwordstretchfactor{4}
\providecommand\BIBentryALTinterwordspacing{\spaceskip=\fontdimen2\font plus
\BIBentryALTinterwordstretchfactor\fontdimen3\font minus \fontdimen4\font\relax}
\providecommand\BIBforeignlanguage[2]{{%
\expandafter\ifx\csname l@#1\endcsname\relax
\typeout{** WARNING: IEEEtran.bst: No hyphenation pattern has been}%
\typeout{** loaded for the language `#1'. Using the pattern for}%
\typeout{** the default language instead.}%
\else
\language=\csname l@#1\endcsname
\fi
#2}}

\bibitem{ref:Multialzu2018Looncopter}
H.~Alzu'bi, I.~Mansour, and O.~Rawashdeh, ``Loon copter: Implementation of a hybrid unmanned aquatic--aerial quadcopter with active buoyancy control,'' \emph{Journal of field Robotics}, vol.~35, no.~5, pp. 764--778, 2018.

\bibitem{ref:Multiravell2018modeling}
D.~A.~M. Ravell, M.~M. Maia, and F.~J. Diez, ``Modeling and control of unmanned aerial/underwater vehicles using hybrid control,'' \emph{Control Engineering Practice}, vol.~76, pp. 112--122, 2018.

\bibitem{ref:fixwei2022lifting}
Z.~Wei, Y.~Teng, X.~Meng, B.~Yao, and L.~Lian, ``Lifting-principle-based design and implementation of fixed-wing unmanned aerial--underwater vehicle,'' \emph{Journal of Field Robotics}, vol.~39, no.~6, pp. 694--711, 2022.

\bibitem{ref:fixweisler2017testing}
W.~Weisler, W.~Stewart, M.~B. Anderson, K.~J. Peters, A.~Gopalarathnam, and M.~Bryant, ``Testing and characterization of a fixed wing cross-domain unmanned vehicle operating in aerial and underwater environments,'' \emph{IEEE Journal of Oceanic Engineering}, vol.~43, no.~4, pp. 969--982, 2017.

\bibitem{ref:TJ-FlyingFishliu2023tj}
X.~Liu, M.~Dou, D.~Huang, S.~Gao, R.~Yan, B.~Wang, J.~Cui, Q.~Ren, L.~Dou, Z.~Gao, \emph{et~al.}, ``Tj-flyingfish: Design and implementation of an aerial-aquatic quadrotor with tiltable propulsion units,'' in \emph{2023 IEEE International Conference on Robotics and Automation (ICRA)}.\hskip 1em plus 0.5em minus 0.4em\relax IEEE, 2023, pp. 7324--7330.

\bibitem{ref:hitchhikingli2022aerial}
L.~Li, S.~Wang, Y.~Zhang, S.~Song, C.~Wang, S.~Tan, W.~Zhao, G.~Wang, W.~Sun, F.~Yang, \emph{et~al.}, ``Aerial-aquatic robots capable of crossing the air-water boundary and hitchhiking on surfaces,'' \emph{Science robotics}, vol.~7, no.~66, p. eabm6695, 2022.

\bibitem{ref:ugJFRlyu2022toward}
C.~Lyu, D.~Lu, C.~Xiong, R.~Hu, Y.~Jin, J.~Wang, Z.~Zeng, and L.~Lian, ``Toward a gliding hybrid aerial underwater vehicle: Design, fabrication, and experiments,'' \emph{Journal of Field Robotics}, vol.~39, no.~5, pp. 543--556, 2022.

\bibitem{ref:ugnezhaoelu2021design}
D.~Lu, C.~Xiong, H.~Zhou, C.~Lyu, R.~Hu, C.~Yu, Z.~Zeng, and L.~Lian, ``Design, fabrication, and characterization of a multimodal hybrid aerial underwater vehicle,'' \emph{Ocean Engineering}, vol. 219, p. 108324, 2021.

\bibitem{ref:wang2024design}
Z.~Wang, Y.~Jiang, Z.~Zou, and Z.~Zhen, ``Design and implementation of a multimodal tilt-rotor unmanned aerial-aquatic vehicle,'' \emph{Journal of Field Robotics}, 2024.

\bibitem{ref:BCFzhu2019tuna}
J.~Zhu, C.~White, D.~K. Wainwright, V.~Di~Santo, G.~V. Lauder, and H.~Bart-Smith, ``Tuna robotics: A high-frequency experimental platform exploring the performance space of swimming fishes,'' \emph{Science Robotics}, vol.~4, no.~34, p. eaax4615, 2019.

\bibitem{ref:BCFwang2020development}
R.~Wang, S.~Wang, Y.~Wang, L.~Cheng, and M.~Tan, ``Development and motion control of biomimetic underwater robots: A survey,'' \emph{IEEE Transactions on Systems, Man, and Cybernetics: Systems}, vol.~52, no.~2, pp. 833--844, 2020.

\bibitem{ref:MPFzhang2022novel}
D.~Zhang, G.~Pan, Y.~Cao, Q.~Huang, and Y.~Cao, ``A novel integrated gliding and flapping propulsion biomimetic manta-ray robot,'' \emph{Journal of Marine Science and Engineering}, vol.~10, no.~7, p. 924, 2022.

\bibitem{ref:MPFarastehfar2019relationship}
S.~Arastehfar, C.-M. Chew, A.~Jalalian, G.~Gunawan, and K.~S. Yeo, ``A relationship between sweep angle of flapping pectoral fins and thrust generation,'' \emph{Journal of mechanisms and robotics}, vol.~11, no.~1, p. 011014, 2019.

\bibitem{ref:flapgu2024bio}
L.~Gu, Y.~Xiang, Z.~Gong, and B.~Tao, ``Bio-inspired wing with bistable morphing airfoils for aquatic-aerial robots,'' \emph{IEEE Robotics and Automation Letters}, 2024.

\bibitem{ref:flaphe2024novel}
J.~He, Y.~Zhang, J.~Feng, S.~Li, Y.~Yuan, P.~Wang, and S.~Han, ``A novel aerial-aquatic unmanned vehicle using flapping wings for underwater propulsion,'' \emph{Biomimetics}, vol.~9, no.~10, p. 581, 2024.

\bibitem{ref:freedomizraelevitz2015novel}
J.~S. Izraelevitz and M.~S. Triantafyllou, ``A novel degree of freedom in flapping wings shows promise for a dual aerial/aquatic vehicle propulsor,'' in \emph{2015 IEEE International Conference on Robotics and Automation (ICRA)}.\hskip 1em plus 0.5em minus 0.4em\relax IEEE, 2015, pp. 5830--5837.

\bibitem{ref:lovell2023attitude}
G.~H. Lovell-Prescod, Z.~Ma, and E.~J. Smeur, ``Attitude control of a tilt-rotor tailsitter micro air vehicle using incremental control,'' in \emph{2023 International Conference on Unmanned Aircraft Systems (ICUAS)}.\hskip 1em plus 0.5em minus 0.4em\relax IEEE, 2023, pp. 842--849.

\bibitem{ref:nelson1998flight}
R.~C. Nelson \emph{et~al.}, \emph{Flight stability and automatic control}.\hskip 1em plus 0.5em minus 0.4em\relax WCB/McGraw Hill New York, 1998, vol.~2.

\bibitem{ref:2017Modeling}
F.~Zhang, X.~Lyu, Y.~Wang, H.~Gu, and Z.~Li, ``Modeling and flight control simulation of a quad rotor tail-sitter vtol uav,'' in \emph{AIAA modeling and simulation technologies conference 2017}, 2017.

\bibitem{ref:fossen2011handbook}
T.~I. Fossen, ``Handbook of marine craft hydrodynamics and motion control,'' \emph{John Willy \& Sons Ltd}, 2011.

\bibitem{ref:etkin1995dynamics}
B.~Etkin and L.~D. Reid, \emph{Dynamics of flight: stability and control}.\hskip 1em plus 0.5em minus 0.4em\relax John Wiley \& Sons, 1995.

\bibitem{ref:liu2020effects}
P.~Liu, S.~Wang, R.~Liu, and Z.~Shang, ``Effects of st and re on propulsive performance of bionic oscillating caudal fin,'' \emph{Ocean Engineering}, vol. 217, p. 107933, 2020.

\bibitem{ref:alam2020dynamics}
M.~M. Alam and Z.~Muhammad, ``Dynamics of flow around a pitching hydrofoil,'' \emph{Journal of Fluids and Structures}, vol.~99, p. 103151, 2020.

\bibitem{ref:crespi2008controlling}
A.~Crespi, D.~Lachat, A.~Pasquier, and A.~J. Ijspeert, ``Controlling swimming and crawling in a fish robot using a central pattern generator,'' \emph{Autonomous Robots}, vol.~25, pp. 3--13, 2008.

\bibitem{ref:cao2015applying}
Y.~Cao, S.~Bi, Y.~Cai, and Y.~Wang, ``Applying central pattern generators to control the robofish with oscillating pectoral fins,'' \emph{Industrial Robot: An International Journal}, vol.~42, no.~5, pp. 392--405, 2015.

\end{thebibliography}

\end{document}